\documentclass[11pt]{article}

\usepackage[margin=1in]{geometry}
\usepackage{amsmath,amssymb,amsfonts}
\usepackage{booktabs}
\usepackage{graphicx}
\usepackage{hyperref}
\usepackage{xcolor}

\newcommand{\method}{MV-GTA}

\newcommand{\simfn}{\operatorname{sim}}

\title{When Does Explicit View Routing Work?\\
A Controlled Study of Multi-View Graph-Text Alignment}
\author{
Xiao Yue \quad Guangzhi Qu \\
Oakland University \\
\texttt{xiaoyue@oakland.edu}, \texttt{gqu@oakland.edu}
}
\date{}

\begin{document}
\maketitle

\begin{abstract}
Graph-text retrieval typically maps a graph and its description to a single
embedding, even when a query concerns only one semantic aspect, such as a class
label or molecular property. Multiple heads can separate these aspects, but a
change in the query head may alter retrieval even when the wrong text is sent
to that head. Such behavior demonstrates architectural channelization, not
necessarily semantic routing. We examine the conditions under which this
distinction can be resolved. Our controlled version of \method{} uses
deterministic, verifiable text segments; isolated text encoders; view-specific
graph heads; and relevance derived from external labels or RDKit descriptors.
Correct routing and per-sample derangements form a causal test of whether
retrieval depends on content. On BBBP and BACE, correct routing improves label
and property nDCG by $0.305$ to $0.685$ over deranged training. The expected
graph head exceeds the best wrong head by $0.303$ to $0.453$. Topology does not
specialize consistently across the two datasets. In a matched three-seed
comparison, one joint model obtains mean topology, label, and property nDCG of
$0.720/1.000/0.877$; three separately trained Single specialists obtain
$0.633/0.976/0.859$. Property paraphrase augmentation also improves
unseen-template nDCG by $0.140$ and $0.147$ over a matched-exposure canonical
control. Consistency and hard-template extensions, however, reduce canonical
retrieval in some settings. The evidence is therefore limited to explicit,
externally grounded label and property routing and observed multi-interface
consolidation. It does not establish free-form routing, consistent three-view
specialization, statistical equivalence to specialists, or superior downstream
prediction.
\end{abstract}

\section{Introduction}

Graph-text retrieval provides a natural-language interface to structured
objects such as molecules, citation networks, and knowledge graphs. Modern
contrastive methods learn this interface by aligning graph and text embeddings
\cite{radford2021clip,edwards2021text2mol,liu2023moleculestm,zhu2024graphclip}.
One-vector representations work well for instance matching, but do not reveal
which semantic aspect accounts for a match. For example, a molecular query may
concern a class outcome, a physicochemical property, or graph structure. These
criteria need not produce the same gallery ranking.

Multiple graph and text heads, one for each semantic view, are a plausible
alternative. Yet a diagonal cross-view retrieval matrix alone does not
demonstrate semantic decomposition. Contrastively aligning each text channel
with one graph channel can make the retrieval subspace depend on the selected
channel, even if that channel receives semantically incorrect text. We refer
to this failure mode as \emph{architectural channelization}: the interface
responds to the selected view, but the association between its content and name
has not been verified.

We ask when a multi-view graph-text model learns content-dependent semantic
routing. The analysis relies on causal controls rather than absolute retrieval
scores alone. Each molecule has deterministic, independently verifiable
topology, label, and property text. The segments are encoded separately and
cannot exchange information before projection. We compare correct routing with
a per-sample derangement in which no segment enters its named channel. During
evaluation, correctly routed and deranged queries search a training gallery.
External labels, graph statistics, or RDKit descriptors define relevance,
independently of the text encoder.

The experiments recover semantic routing for label and property, but not for
all three views. On BBBP and BACE, label and property retrieval responds to the
intended content and favors the expected graph head. BACE topology is the
exception: its expected head does not outperform the best wrong head. An
externally grounded property objective improves property retrieval, although
conventional molecular fingerprint baselines remain better for downstream logD
prediction. We accordingly treat routing as a retrieval and diagnostic
capability, without claiming predictive superiority. A matched specialist
control provides a separate descriptive result. In the three evaluated seeds,
one joint model offers all three query interfaces with higher mean aggregate
retrieval scores than three independently trained specialists. This observation
does not establish statistical equivalence, non-inferiority, or stable per-view
superiority.

We also test routing under changes in linguistic form. Label routing transfers
with little change to held-out paraphrases. Property routing remains sensitive
to content and facts, but retrieval quality declines under unseen syntax.
Matched-budget paraphrase augmentation recovers much of this loss on both
datasets. Consistency training and hard paraphrases expose a trade-off:
robustness may improve while canonical retrieval declines. Average paraphrase
gains do not capture this cost.

The study makes the following contributions:
\begin{itemize}
    \item A correct-versus-deranged control with external relevance
    distinguishes semantic routing from architectural channelization.
    \item Strictly supervised routing is established for label and property on
    two molecular datasets; topology specialization remains dataset-dependent.
    \item In three evaluated seeds, one joint model has higher mean aggregate
    topology, label, and property retrieval scores than separate specialists.
    The comparison is descriptive and measures consolidation, not statistical
    equivalence, non-inferiority, or per-view superiority.
    \item A controlled paraphrase analysis evaluates robustness gains alongside
    canonical retrieval, content sensitivity, head specificity, and
    fact following.
\end{itemize}

\section{Related work}

\paragraph{Graph representation learning.}
Message-passing networks and graph transformers combine local structure with
node and edge attributes \cite{gilmer2017mpnn,xu2019gin,rampasek2022gps}.
Graph contrastive methods further learn representations from augmented views or
mutual-information objectives \cite{velickovic2019dgi,sun2020infograph,you2020graphcl}.
These methods generally optimize one graph representation. We do not assume
that the semantic factors are statistically independent. Our question is
whether explicitly named retrieval channels become dependent on their intended
content.

\paragraph{Graph-language alignment.}
Text2Mol, MoleculeSTM, and GraphCLIP align molecular or graph representations
with language through contrastive learning
\cite{edwards2021text2mol,liu2023moleculestm,zhu2024graphclip}. Their goal is
to learn a useful shared space. Our concern is diagnostic: when a model exposes
several cross-modal spaces as semantic views, what evidence shows that the view
names correspond to content rather than head indices?

\paragraph{Supervised contrastive and structured targets.}
Supervised contrastive learning uses known semantic relations to define
positives beyond paired instances \cite{khosla2020supcon}. Graph kernels can
define structured graph similarity \cite{shervashidze2011wl}, while molecular
descriptors provide a chemistry-grounded similarity signal. We compute the
latter with RDKit 2026.03.3 \cite{rdkit2026}. Labels and RDKit descriptors
serve as external anchors and define evaluation relevance independently of the
text encoder, avoiding the circular use of one semantic source for both
representation and evaluation.

\section{Method}

\subsection{Problem setting}

Let $\mathcal{D}=\{(G_i,T_i)\}_{i=1}^{N}$ contain a graph $G_i$ and an explicitly
segmented description:
\[
T_i=(T_i^{\mathrm{top}},T_i^{\mathrm{label}},T_i^{\mathrm{prop}}).
\]
The setting is strongly supervised: segment identities are given rather than
inferred. For each view $v$, the model produces a graph embedding $g_i^v$ and a
text embedding $t_i^v$. A query $t_i^v$ ranks a train-gallery graph $j$ using
$\simfn(t_i^v,g_j^v)$. Evaluation is cross-split; the query molecule itself is
never a gallery item.

We define content-dependent routing by two requirements. Correct content must
retrieve external semantic neighbors more accurately than deranged content,
and the expected graph head must outperform the best wrong head for the same
query. The latter requirement rules out success caused by every head encoding
the same useful molecular information.

\subsection{Strict explicit text routing}

Earlier implementations encoded the full description before pooling by view.
Contextual attention could then leak information across segments despite a
one-hot segment mask. Applying a softmax to this mask also assigned nonzero mass
to unintended views. The strict path removes both sources of ambiguity.

Each segment is independently encoded by a frozen language encoder $E_T$ and a
view-specific projection $P_T^v$:
\begin{equation}
    t_i^v = \frac{P_T^v(E_T(T_i^v))}
    {\|P_T^v(E_T(T_i^v))\|_2}.
\end{equation}
The model uses neither full-text self-attention nor soft token assignment or
cross-segment pooling. It tests semantic routing with trusted view tags, not
automatic discovery from free-form text.

\subsection{View-specific graph representations}

A graph backbone encodes the molecule and view-specific graph pathways produce
normalized embeddings:
\begin{equation}
    g_i^v = \frac{P_G^v(E_G^v(G_i))}
    {\|P_G^v(E_G^v(G_i))\|_2}.
\end{equation}
Each independent pathway gives its query view a separate retrieval space. The
pathways alone do not demonstrate specialization. The routing controls test
whether each space acquires its intended meaning.

\subsection{Cross-modal alignment and external anchors}

The base objective applies a symmetric contrastive loss separately to each
view. For the graph-to-text direction,
\begin{equation}
 \mathcal{L}_{g\rightarrow t}^{v}
 =-\frac{1}{B}\sum_i
 \log\frac{\exp(\simfn(g_i^v,t_i^v)/\tau)}
 {\sum_j\exp(\simfn(g_i^v,t_j^v)/\tau)},
\end{equation}
with an analogous text-to-graph term. The hard-alignment objective averages the
two directions and the three views.

In the externally grounded M2 variant, label equality and RDKit descriptor
similarity define soft target neighborhoods for label and property. These
targets modify the positive distribution without asking the text encoder which
gallery molecules are semantically related. Graph-derived statistics provide
the topology targets. We report topology as a diagnostic because it does not
consistently meet the specialization criterion.

\subsection{Causal routing controls}

Correct and deranged routing differ only in the segment-to-channel mapping. A
derangement $\pi_i$ is independently sampled for each molecule such that
$\pi_i(v)\neq v$ for all three views. The deranged condition feeds
$T_i^{\pi_i(v)}$ into channel $v$. Because the mapping changes per sample and
has no fixed points, the model cannot recover the named correspondence from a
global permutation.

We use three complementary comparisons. \emph{Content sensitivity} evaluates
one correctly trained model with correct versus deranged test content.
\emph{Training mapping} compares correct-trained and deranged-trained models
under correct test content. \emph{End-to-end routing} compares each model under
its native mapping. Semantic routing requires positive effects for label and
property together with expected-head specificity.

\subsection{Paraphrase robustness extensions}

Canonical descriptions list five property facts: LogP, TPSA, hydrogen-bond
donors, hydrogen-bond acceptors, and formal charge. Paraphrase templates change
surface form while code fills the same structured facts. Train and held-out
template families are disjoint. Deranged-donor and minimal-change queries test
whether a model responds to content instead of only tolerating surface changes.

The augmentation variant adds ordinary training paraphrases. A consistency
extension adds a one-way loss from paraphrased to canonical property text,
\begin{equation}
 \mathcal{L}_{\mathrm{cons}}
 =1-\cos(t_i^{\mathrm{prop,para}},
 \operatorname{stopgrad}(t_i^{\mathrm{prop,canon}})).
\end{equation}
A hard-paraphrase experiment compares an additional difficult syntax pool with
an ordinary pool of equal size. We evaluate these extensions under canonical
non-inferiority and fact-following constraints. The graph architecture and
strict routing definition remain unchanged.

\section{Experiments}

\subsection{Experimental setup}

\paragraph{Datasets and text construction.}

We use scaffold splits of BBBP and BACE, with 600 molecules per dataset
($480/60/60$ train/validation/test) and 1,200 molecules in total in the
controlled routing benchmark. Topology text reports RDKit-derived ring,
aromatic-ring, and rotatable-bond facts. Label text explicitly states the sample
outcome and is therefore an upper-bound semantic-routing protocol, not a
leakage-free classification input. Property text reports LogP, TPSA, HBD, HBA,
and formal charge. All canonical facts are generated deterministically.

For natural-language stress tests, a language model creates template skeletons
only; structured facts are filled by code. Automatic checks enforce complete
slot coverage, forbid dataset/task/label terms in property text, and verify
train/evaluation family separation. All 1,200 Phase 2F-P records pass these
checks. No chemistry-domain expert audit was available, so we report automatic
validation rather than expert validation.

\paragraph{Metrics and statistics.}

The main metric is test-to-train nDCG@10 under view-specific external relevance.
We also report content sensitivity, expected-head specificity,
canonical-to-paraphrase retention, and five-field fact following. Each formal
controlled-routing and paraphrase experiment uses seeds $0,1,2$, 50 epochs,
batch size 32, and learning rate $2\times10^{-5}$. Confidence intervals use
5,000 hierarchical bootstrap resamples over seeds and queries. Method
comparisons are paired by seed and query.
The specialist-parity study reports mean and standard deviation over the three
training seeds, together with per-seed paired directions. Query-level
bootstraps within a trained seed are treated as diagnostics rather than as a
substitute for additional independent training seeds or a formal equivalence
test.

\paragraph{Guardrails.}

An improvement on paraphrased queries is insufficient by itself. The
pre-specified checks require canonical non-inferiority, positive
paraphrase-versus-deranged content sensitivity, positive expected-head
specificity, at least 80\% held-out retention, and no material degradation on
any of the five property fields. Together, these checks reject models that
raise the average robustness score by ignoring facts or sacrificing the
canonical interface.

\subsection{Strict routing recovers label and property semantics}

Table \ref{tab:m2-content} reports positive correct-minus-deranged effects for
every dataset and view. The three-view criterion still fails because BACE
topology does not favor its expected graph head. Label and property specificity
is positive on both datasets, with confidence intervals that exclude zero.

\begin{table}[t]
\caption{Strict M2 routing. Content is correct minus deranged test content;
specificity is expected graph head minus the best wrong head. Values are nDCG
differences with 95\% hierarchical bootstrap confidence intervals.}
\label{tab:m2-content}
\centering
\small
\begin{tabular}{llcc}
\toprule
Dataset & View & Content sensitivity $\uparrow$ & Head specificity $\uparrow$ \\
\midrule
BACE & Topology & $+0.256$ $[+0.191,+0.314]$ & $+0.000$ $[-0.035,+0.033]$ \\
BACE & Label    & $+0.426$ $[+0.361,+0.493]$ & $+0.453$ $[+0.376,+0.515]$ \\
BACE & Property & $+0.617$ $[+0.576,+0.658]$ & $+0.333$ $[+0.305,+0.361]$ \\
BBBP & Topology & $+0.306$ $[+0.213,+0.419]$ & $+0.115$ $[+0.040,+0.209]$ \\
BBBP & Label    & $+0.471$ $[+0.402,+0.540]$ & $+0.408$ $[+0.361,+0.457]$ \\
BBBP & Property & $+0.447$ $[+0.375,+0.533]$ & $+0.303$ $[+0.260,+0.339]$ \\
\bottomrule
\end{tabular}
\end{table}

\begin{figure}[t]
\centering
\includegraphics[width=0.92\textwidth]{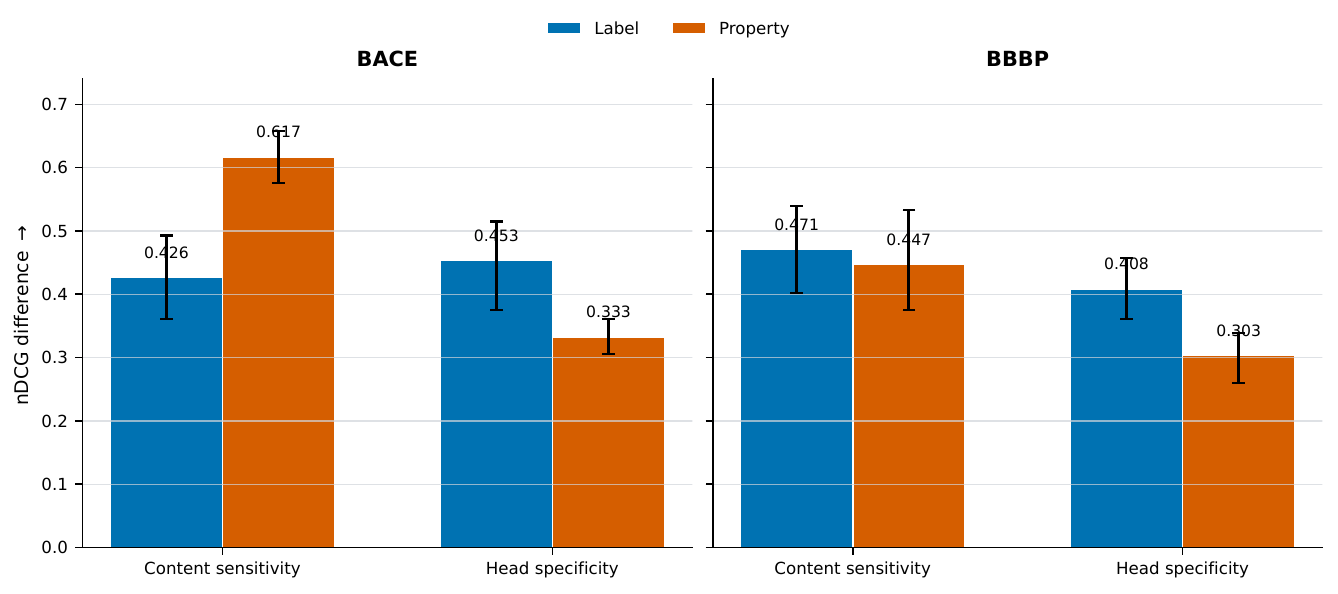}
\caption{Strict M2 label/property routing on BACE and BBBP. Content sensitivity
is correct minus deranged test content; head specificity is the expected graph
head minus the best wrong head. Bars show paired mean nDCG differences and
error bars show 95\% hierarchical seed/query bootstrap intervals.}
\label{fig:strict-routing-results}
\end{figure}

Figure~\ref{fig:strict-routing-results} isolates the four supported
label/property comparisons. All content-sensitivity and expected-head effects
are positive with confidence intervals excluding zero on both datasets.

The training mapping matters in addition to test-time content.
Correct-trained minus deranged-trained nDCG is $+0.635/+0.305$ for BACE/BBBP
label and $+0.647/+0.685$ for property; all four confidence intervals exclude
zero. M2's external property anchor improves expected-head property
nDCG over hard-CLIP M1 by $+0.056$ on BACE and $+0.080$ on BBBP. The result is
specific to property: external grounding improves its retrieval, while
topology remains unresolved.

\subsection{Observed joint-model means relative to specialists}

To determine whether a joint interface retains the quality of dedicated
models, we compare one strict Full model with three Single specialists. The
Full model is trained jointly on all views; each specialist is trained and
evaluated on topology, label, or property text alone. The split, deterministic
text, graph input, optimizer, and per-view CLIP weight are matched. The topology
specialist uses the same raw structural graph input and projection path as the
Full topology view.

The joint Full configuration has a higher three-seed mean than each specialist
in Table~\ref{tab:specialist-parity}. We use this as descriptive evidence of
\emph{functional consolidation}: in the evaluated runs, one model provides
three independently callable interfaces with higher mean aggregate retrieval
scores than separate view-only models. This comparison does not establish
statistical equivalence, non-inferiority, or a uniformly stronger
representation. Full-minus-specialist differences across seeds are
$-0.075/+0.241/+0.096$ for topology,
$+0.072/0.000/0.000$ for label, and $+0.007/+0.027/+0.019$ for property.
Topology has one unfavorable seed, and the label comparison lies near the
metric ceiling.

\begin{table}[t]
\caption{Specialist-parity study. Values are pooled BACE/BBBP test-to-train
nDCG@10, reported as mean $\pm$ standard deviation over three training seeds.
All Full entries come from the same jointly trained configuration; each Single
specialist is trained separately on only its named view. $\Delta$ is Full minus
Single.}
\label{tab:specialist-parity}
\centering
\small
\begin{tabular}{lccc}
\toprule
Query interface & Single specialist & Joint Full & $\Delta$ \\
\midrule
Topology & $0.633 \pm 0.069$ & $\mathbf{0.720 \pm 0.062}$ & $+0.087$ \\
Label    & $0.976 \pm 0.034$ & $\mathbf{1.000 \pm 0.000}$ & $+0.024$ \\
Property & $0.859 \pm 0.006$ & $\mathbf{0.877 \pm 0.002}$ & $+0.018$ \\
\bottomrule
\end{tabular}
\end{table}

The specialist comparison and cross-head specialization are distinct. A
topology interface can retrieve externally defined topology neighbors with a
higher observed mean than its specialist without outperforming every wrong
Full-model head on BACE. Thus Table~\ref{tab:specialist-parity} provides
descriptive evidence of multi-interface consolidation, but the three-view
specialization criterion in Table~\ref{tab:m2-content} remains unsatisfied.

\subsection{Why the strict control is necessary}

In the historical soft path, changing the query channel produced a diagonal
retrieval pattern, but per-sample text derangement did not reliably reduce the
label/property gap. Correct minus deranged confidence intervals crossed zero,
whereas removing distinct text heads collapsed the gap. The original
observation can therefore be attributed mainly to architectural channelization.
Under strict segment isolation and external relevance, incorrect content
measurably harms retrieval.

\subsection{Natural paraphrases expose a property weakness}

Table \ref{tab:paraphrase-base} evaluates the strict model on held-out natural
templates. Label routing changes little. Property retains 86.5\% of canonical
nDCG and remains strongly content-sensitive. On BACE, however, sentence form
affects representation distance more than a change to one fact. Property
routing is therefore present but sensitive to linguistic form.

\begin{table}[t]
\caption{Canonical-to-held-out paraphrase transfer for the strict M2 model.
Content sensitivity is paraphrase minus a same-head deranged donor.}
\label{tab:paraphrase-base}
\centering
\small
\begin{tabular}{llccc}
\toprule
Dataset & View & Canonical & Paraphrase & Retention \\
\midrule
BACE & Label    & 1.000 & 0.998 & 99.8\% \\
BACE & Property & 0.931 & 0.805 & 86.5\% \\
BBBP & Label    & 1.000 & 0.997 & 99.7\% \\
BBBP & Property & 0.926 & 0.801 & 86.5\% \\
\bottomrule
\end{tabular}
\end{table}

\subsection{Paraphrase augmentation helps, with guardrail trade-offs}

Ordinary property paraphrase augmentation gives the largest practical
improvement. Relative to a matched-exposure canonical control on the same
held-out set, BACE nDCG increases by $+0.140$ with a 95\% confidence interval
of $[+0.099,+0.185]$. On BBBP, the increase is $+0.147$, with an interval of
$[+0.082,+0.211]$. Retention rises to 88.7\% and 87.8\%, while content
sensitivity, head specificity, and fact following remain positive. The strict
canonical guardrail narrowly misses on BBBP: its non-inferiority lower bound is
$-0.021$ against a pre-specified $-0.020$ threshold.

Consistency reduces sentence-form displacement and closes the BBBP canonical
guardrail, but shifts the trade-off to BACE. Relative to the same-run matched
canonical control, the main consistency setting changes canonical nDCG by
$-0.013$ $[-0.027,-0.001]$ on BACE and $+0.008$
$[-0.003,+0.020]$ on BBBP. Held-out gains remain large, and all five property
fields remain protected. Consistency may therefore be useful in practice, but
it does not satisfy the criterion on both datasets.

The consistency evidence is also provisional because a pre-specified
cross-run drift check failed. With code, data, template, and cache fingerprints
held fixed, an independent $\lambda=0$ reproduction differed from the Phase 2D
held-out paraphrase nDCG by $0.034$ on BACE and $0.022$ on BBBP, both above the
locked $0.01$ tripwire. The within-run paired comparisons above remain
traceable, but the drift limits claims about run-level stability.

\begin{table}[t]
\caption{Summary of paraphrase interventions. Positive held-out effects do not
override failed canonical guardrails. MC denotes matched canonical exposure.}
\label{tab:robustness-summary}
\centering
\small
\begin{tabular}{llcc}
\toprule
Intervention & Comparison & BACE $\Delta$ nDCG & BBBP $\Delta$ nDCG \\
\midrule
Ordinary augmentation & Held-out vs. MC & $+0.140$ & $+0.147$ \\
Consistency $\lambda=.05$ & Canonical vs. MC & $-0.013$ & $+0.008$ \\
Consistency $\lambda=.05$ & Held-out vs. MC & $+0.108$ & $+0.152$ \\
Hard extra pool & Fresh hard vs. base & $-0.015$ & $-0.022$ \\
Hard extra pool & Canonical vs. base & $-0.026$ & $-0.034$ \\
\bottomrule
\end{tabular}
\end{table}

\begin{figure}[t]
\centering
\includegraphics[width=0.92\textwidth]{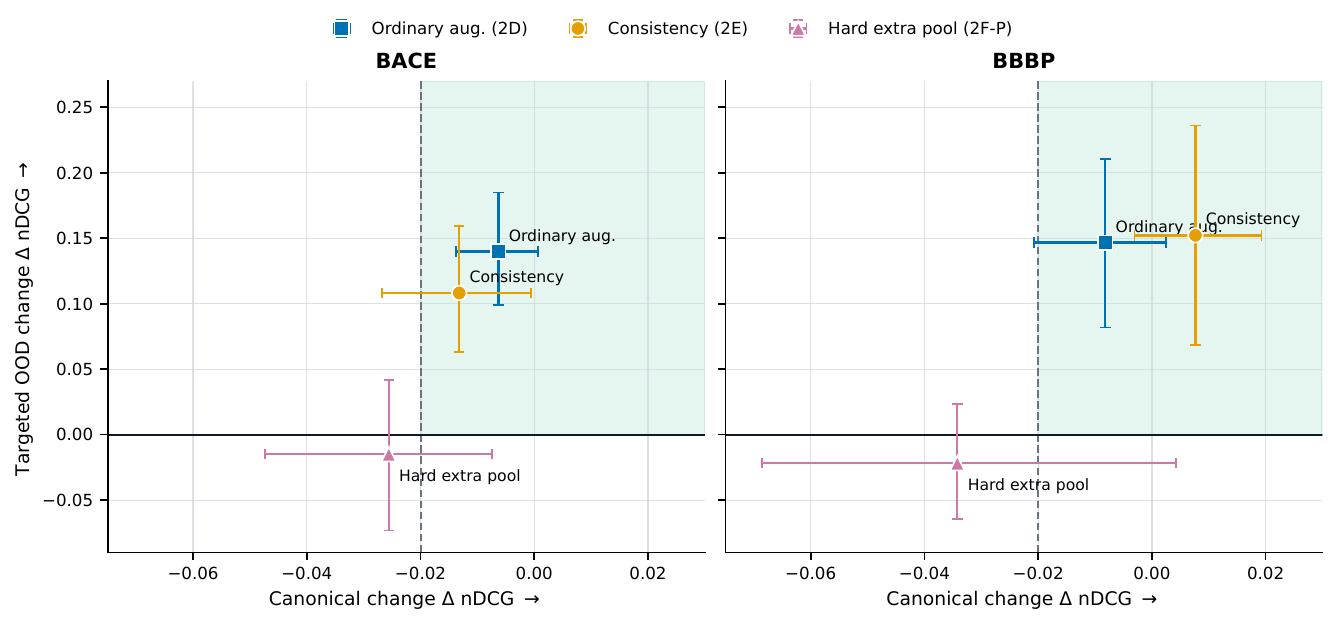}
\caption{Canonical versus OOD trade-off for paraphrase interventions. Horizontal and
vertical error bars are 95\% hierarchical bootstrap intervals. The shaded
upper-right region satisfies canonical change $\geq-0.02$ and targeted OOD
change $>0$. The OOD axis uses held-out ordinary paraphrases for Phase 2D/2E
and the separately locked fresh-hard set for Phase 2F-P; it should be read as
an intervention-relative target, not as one shared test set.}
\label{fig:robustness-tradeoff}
\end{figure}

Figure~\ref{fig:robustness-tradeoff} summarizes the guardrail trade-off.
Ordinary augmentation and consistency improve their targeted held-out queries,
but only some dataset-method pairs remain on the protected side of the
canonical threshold. The hard extra pool is below zero on its fresh-hard target
as well as outside the canonical protected region.

\subsection{Hard syntax is better than matched ordinary expansion, but not the base}

The hard-paraphrase pool isolates data type from additional exposure. Against
an equal-size ordinary extra pool, hard syntax improves pooled fresh-hard nDCG
by $+0.039$ $[+0.017,+0.060]$. The improvement is concentrated in one fresh
template family; it is clear on BBBP but uncertain on BACE. The hard pool also
does not beat the simpler consistency base: pooled fresh-hard
performance changes by $-0.018$ $[-0.066,+0.028]$, and canonical nDCG falls by
$0.026/0.034$. Given the same additional exposure budget, hard syntax
outperforms ordinary syntax, but the second pool provides no net benefit over
the base.

\subsection{Independent property and downstream baselines narrow the claim}

An independent analysis using RDKit descriptors finds limited property information
beyond the label head after controlling for molecular size, but the property
head does not consistently beat topology and is not unique to the full
multi-view model. ECFP+descriptor Random Forest baselines obtain
RMSE $0.803$ on random-split Lipo and $0.923$ on scaffold Lipo, compared with
approximately $0.950$ and $1.137$ for the MV-GTA property readout. The comparison
does not support downstream regression superiority or a property geometry
unique to the full model.

\section{Discussion}

A controllable retrieval interface does not by itself imply semantic
decomposition. Independent heads are sufficient to change retrieval when a
user switches channels. Semantic routing also requires correct content and the
correct training mapping to improve an externally defined metric. Explicit
label and property routing meets this requirement in our controlled setting.

Robustness must be evaluated under several constraints. Paraphrase augmentation
produces a large held-out gain across the three evaluated seeds, but neither
consistency training nor difficult-template expansion preserves the canonical
interface on both datasets. The positive content-sensitivity and fact-following
results indicate that the model has not simply ignored the properties. Instead,
the results suggest a representation and optimization trade-off between
familiar and shifted language forms.

Taken together, the experiments characterize \method{} as a controlled routing
framework rather than a general performance method. It provides an explicit
semantic query interface and a protocol for testing that interface. In the
specialist comparison, one joint configuration has higher aggregate three-seed
mean retrieval scores across topology, label, and property. This is descriptive
evidence of interface consolidation, not a demonstrated interface-level
advantage: no equivalence or non-inferiority margin was pre-specified. It does
not imply that multi-view training improves every representation or outperforms
strong single-view and graph-only methods on downstream prediction.

\section{Limitations and future work}

The protocol relies on strong supervision. Text segments and view identities
are supplied explicitly, label text contains the sample outcome, and property
relevance is defined using RDKit descriptors. The results establish a learnable
upper-bound routing mechanism, but not automatic routing of arbitrary language
or leakage-free label prediction.

The evaluation is also intentionally narrow. Main controlled experiments use
BBBP and BACE with three training seeds, although hierarchical bootstrap also
resamples queries. Topology specialization is inconsistent across these two
datasets, the topology parity direction changes in one seed, and label parity
is ceiling-limited. The specialist study therefore provides descriptive
three-seed mean comparisons rather than statistical equivalence or
non-inferiority evidence. It also does not measure parameter storage,
wall-clock training, peak memory, or throughput against three separately
deployed specialists. No chemistry-domain expert was available for paraphrase
auditing;
automatic slot, leakage, and family-isolation checks passed, but semantic audit
claims are limited accordingly.

The consistency experiments also failed a pre-specified cross-run drift
tripwire on held-out paraphrase retrieval despite matching recorded
fingerprints, so their run-level stability remains provisional.

A larger study could use size-controlled Lipo pairs to test whether molecular
size and topology explain continuous-target retrieval. ESOL would provide a
transfer test on another regression dataset, and the selected configurations
could be evaluated with five rather than three seeds. Because these experiments
have not been run, they do not support any claim in this paper.

\section{Conclusion}

We studied explicit multi-view graph-text routing under strict segment
isolation, external semantic relevance, and per-sample derangements. Label and
property routing is learnable on BBBP and BACE, whereas topology does not meet
the cross-dataset specialization criterion. In the specialist comparison, one
joint model has higher observed three-seed mean aggregate retrieval across all
three interfaces, although this does not establish equivalence,
non-inferiority, or stable per-view superiority. Paraphrase
augmentation substantially improves robustness, while consistency training and
hard syntax reduce canonical retrieval in some settings. The evidence supports
explicit, externally grounded semantic routing and multi-interface
consolidation. It does not support free-form view discovery, consistent
three-view specialization, statistical equivalence to specialists, or superior
downstream prediction.

\bibliographystyle{plain}
\bibliography{references}

\appendix

\section{Reproducibility details}

\paragraph{Data and retrieval direction.}
The controlled BBBP and BACE benchmarks contain 600 molecules per dataset and
use locked scaffold splits. Each formal statistic is computed from 60 test
queries per dataset and seed. The gallery contains 480 training molecules per
dataset, so
the evaluated direction is always test text to train graph. Within-split and
self-match retrieval are retained only as diagnostics and are not used for a
paper claim.

\paragraph{Training configuration.}
All main controlled-routing and paraphrase runs use seeds 0, 1, and 2, with 50
epochs, a batch size of 32, and a learning rate of $2\times10^{-5}$. Paired comparisons keep the
dataset split, graph inputs, text families, model architecture, and optimizer
configuration fixed unless that component is the declared intervention.
Matched-exposure controls repeat canonical text to equalize the number of text
examples and updates without introducing new syntax.

\paragraph{Statistical unit and traceability.}
Confidence intervals are produced by 5,000 hierarchical bootstrap samples:
seeds are resampled first and paired queries are resampled within each selected
seed. We report paired means, confidence intervals, per-seed directions, and
family- or field-level diagnostics where applicable. Formal outputs retain
per-query scores together with split, template, canonical-text, data, and
configuration hashes, allowing every table entry to be traced back to its
locked run.

\section{Complete claim-evidence boundary}

Table~\ref{tab:claim-boundary} specifies the scope of each claim. A claim is
``supported'' if its full operational definition is satisfied on both BBBP and
BACE under the paired hierarchical bootstrap protocol. ``Supported, bounded''
indicates a positive practical effect that fails a pre-specified protection
criterion. ``Not evaluated'' means that the required setting falls outside the
current explicit-routing protocol; it is not a negative result.

\begin{table}[p]
\caption{Operational meaning and evidence boundary of the paper's claims. A
claim is not treated as supported when only one component of its operational
definition passes.}
\label{tab:claim-boundary}
\centering
\footnotesize
\setlength{\tabcolsep}{3pt}
\renewcommand{\arraystretch}{1.18}
\begin{tabular}{p{0.21\textwidth}p{0.32\textwidth}p{0.13\textwidth}p{0.28\textwidth}}
\toprule
Claim & Operational meaning & Status & Evidence or reason \\
\midrule
Explicit label/property routing is content-dependent &
The named text channel must depend on the intended semantic facts, rather than
only on its head index. Correct content and correct training mapping must beat
zero-fixed-point derangements, and the expected graph head must beat the best
wrong head on both datasets. &
\textbf{Supported} &
All label/property content-sensitivity and training-mapping effects are
positive with CIs excluding zero. Expected-head gains range from
$+0.303$ to $+0.453$. \\
\addlinespace
External property anchors improve property retrieval &
Descriptor-derived soft neighborhoods in M2 must improve the property channel
over hard instance-only alignment in M1. This is a selective property claim,
not a claim that external anchors improve every view. &
\textbf{Supported} &
M2 minus M1 expected-head property nDCG is $+0.056$ on BACE and $+0.080$ on BBBP;
both CIs exclude zero. Topology does not show the same improvement. \\
\addlinespace
One joint model has higher observed mean retrieval &
The same jointly trained Full configuration must expose topology, label, and
property interfaces whose observed three-seed mean nDCG is higher than that of
separately trained, input-matched Single specialists. This is a descriptive
consolidation result, not a claim of positive transfer, equivalence, or
non-inferiority. &
\textbf{Observed, bounded} &
Full versus Single specialist nDCG is $0.720/0.633$ for topology,
$1.000/0.976$ for label, and $0.877/0.859$ for property. Full is higher in all
three means, but topology has one unfavorable seed, label is ceiling-limited,
and no equivalence or non-inferiority margin was pre-specified. \\
\addlinespace
Ordinary paraphrase augmentation improves unseen syntax &
With matched text exposure and optimizer budget, training on varied property
expressions must outperform repeated canonical text on disjoint held-out
template families while retaining content, head, and fact sensitivity. &
\textbf{Supported, bounded} &
Held-out gains are $+0.140/+0.147$. Content, specificity, and five-field fact
checks pass, but the BBBP canonical non-inferiority lower bound is $-0.021$,
just outside the $-0.020$ gate. \\
\addlinespace
Consistency removes the robustness-canonical trade-off &
Canonical-to-paraphrase consistency must retain the held-out gain and satisfy the
canonical non-inferiority gate simultaneously on both datasets. Improvement on
only one dataset is insufficient. &
\textbf{Not supported} &
Held-out gains remain $+0.108/+0.152$, and BBBP canonical protection closes,
but BACE canonical changes by $-0.013$ with CI lower bound $-0.027$. \\
\addlinespace
All three views specialize across datasets &
Topology, label, and property queries must each prefer their named graph head
over the best wrong head on both BBBP and BACE. Content sensitivity alone does
not establish head specialization. &
\textbf{Not supported} &
Label and property pass all four dataset-view cells, but BACE topology
specificity is $+0.000$ $[-0.035,+0.033]$. \\
\addlinespace
Free-form or learned routing works &
The model must infer view assignments from unsegmented, arbitrary language
without explicit view tags or deterministic segment boundaries. Robustness to
paraphrases inside known property slots is not equivalent to this ability. &
\textbf{Not evaluated} &
The strict path receives explicit topology, label, and property segments and
uses separate encodings. No learned splitter or untagged routing benchmark is
used. \\
\bottomrule
\end{tabular}
\end{table}

\clearpage

\begin{table}[p]
\caption{Remaining claim boundaries, continuing Table~\ref{tab:claim-boundary}.}
\label{tab:claim-boundary-continuation}
\centering
\footnotesize
\setlength{\tabcolsep}{3pt}
\renewcommand{\arraystretch}{1.18}
\begin{tabular}{p{0.21\textwidth}p{0.32\textwidth}p{0.13\textwidth}p{0.28\textwidth}}
\toprule
Claim & Operational meaning & Status & Evidence or reason \\
\midrule
MV-GTA improves downstream prediction &
The full multi-view model must outperform comparable single/no-view models and
standard graph-only predictors on supervised classification or regression. A
readable semantic channel alone is insufficient. &
\textbf{Not supported} &
The full model has no stable downstream advantage. On Lipo, ECFP+descriptor RF
obtains random/scaffold RMSE $0.803/0.923$, better than MV-GTA's approximately
$0.950/1.137$. \\
\addlinespace
Hard-paraphrase expansion has a net benefit &
The hard pool must beat both an equal-size ordinary extra pool and the simpler
base without a second pool on fresh-hard queries, while preserving canonical
retrieval. Beating only the matched ordinary pool is a local advantage. &
\textbf{Not supported} &
Hard beats matched ordinary by pooled $+0.039$, but hard minus base is
$-0.018$ and canonical changes by $-0.026/-0.034$ on BACE/BBBP. \\
\bottomrule
\end{tabular}
\end{table}

\section{Automatic paraphrase validation}

\paragraph{Fact completeness and provenance.}
Every property template contains one slot each for LogP, TPSA, HBD, HBA, and
formal charge. The template supplies the linguistic structure, and code inserts
the five values from structured RDKit records. A missing, duplicated, or
unparsable slot causes validation to fail; the sample is not removed after
evaluation.

\paragraph{Leakage and family isolation.}
Validation rejects dataset, task, and class-label terms in property text. It
also checks that training, development, and fresh held-out template families
share neither a template string nor a declared family identity. Deranged
queries use all five facts from another molecule in the same dataset and head;
minimal-change queries alter one declared field. These controls distinguish
responses to content from template recognition.

\paragraph{Audit outcome and limitation.}
All 1,200 instantiated Phase 2F-P audit records pass the automatic checks for
slot completeness, forbidden terms, and family isolation. A chemistry expert
did not assess linguistic naturalness or domain plausibility. The fixed audit
status is ``automatic validation passed; no
domain-expert validation'' and do not describe the text as
``expert-validated.''

\section{Pre-specified non-claim-bearing extensions}

This section records protocols for optional extensions; it reports no results
and is not evidence for the paper's claims.

\paragraph{Size-controlled Lipo.}
Query and gallery candidates would first be matched or stratified by heavy-atom
count and topology statistics. The evaluation would then measure
target-distance triplet accuracy and partial Spearman correlation after
controlling for size. This test would determine whether molecular size or
topology accounts for the apparent continuous property geometry. It is required
only for a stronger claim about Lipo property specialization.

\paragraph{ESOL transfer.}
ESOL would reuse the explicit three-segment construction, scaffold split, and
the same property-, label-, single-, and no-view comparisons. The first stage
would use three seeds. Seeds 3 and 4 would be added only if the property-minus-
label direction is consistent and practically meaningful; otherwise ESOL would
be reported as a boundary rather than used to tune the method.

\paragraph{Five-seed confirmation.}
A confirmation run would add seeds 3 and 4 only for the final configurations
selected before observing those seeds. Failed Phase 2E/2F-P variants would not
be expanded merely to search for a favorable average. This extension would
strengthen run-level stability but would not change the current query-level
bootstrap evidence.

\section*{Disclosure of Generative AI Use}

Generative AI tools assisted with code generation and debugging, experimental
design, and the drafting and editing of parts of the manuscript. 

\end{document}